\documentclass[lettersize,journal]{IEEEtran}
\usepackage[cmex10]{amsmath}
\usepackage{amsfonts}
\usepackage{ifpdf}
\usepackage{algorithmic}
\usepackage{algorithm}

\usepackage{array}
\usepackage[caption=false,font=normalsize,labelfont=sf,textfont=sf]{subfig}
\usepackage{textcomp}
\usepackage{stfloats}
\usepackage{url}
\usepackage{verbatim}
\usepackage[pdftex]{graphicx}
\usepackage{cite}
\newcommand{\RR}{\mathbb{R}}
\newcommand{\ZZ}{\mathbb{Z}}

\newcommand{\rd}{\mathrm{d}}

\begin{document}

\title{Timing-Based Backpropagation in Spiking Neural Networks Without Single-Spike Restrictions}

\author{Kakei Yamamoto\thanks{Kakei Yamamoto is with the University of Tokyo, Tokyo, Japan (e-mail: kakei@g.ecc.u-tokyo.ac.jp).}, Yusuke Sakemi\thanks{Yusuke Sakemi is with Research Center for Mathematical Engineering, Chiba Institute of Technology, Narashino, Japan.}, Kazuyuki Aihara\thanks{Kazuyuki Aihara is with the International Research Center for Neurointelligence (WPI-IRCN), The University of Tokyo Institutes for Advanced Study, The University of Tokyo, Tokyo, Japan.}}

\maketitle

\begin{abstract}

We propose a novel backpropagation algorithm for training spiking neural networks (SNNs) that encodes information in the relative multiple spike timing of individual neurons without single-spike restrictions. The proposed algorithm inherits the advantages of conventional timing-based methods in that it computes accurate gradients with respect to spike timing, which promotes ideal temporal coding. Unlike conventional methods where each neuron fires at most once, the proposed algorithm allows each neuron to fire multiple times. This extension naturally improves the computational capacity of SNNs. Our SNN model outperformed comparable SNN models and achieved as high accuracy as non-convolutional artificial neural networks. The spike count property of our networks was altered depending on the time constant of the postsynaptic current and the membrane potential. Moreover, we found that there existed the optimal time constant with the maximum test accuracy. That was not seen in conventional SNNs with single-spike restrictions on time-to-fast-spike (TTFS) coding. This result demonstrates the computational properties of SNNs that biologically encode information into the multi-spike timing of individual neurons. Our code would be publicly available. 

\end{abstract}


\section{Introduction}

\IEEEPARstart{H}{ow} does the human brain acquire intelligence? Neural systems in the brain have various functions by forming very complex networks, and much remains a mystery. One approach to this question is to use mathematical models that represent the dynamics of biological neurons from a microscopic chemical viewpoint. You can simulate and analyze spiking phenomena, and investigate the functions of the biological neural system that the model mimics. These models, which originated in the Hodgkin-Huxley model \cite{Hodgkin1952} and the FitzHugh-Nagumo model \cite{FitzHugh1961, Nagumo1962}, can reproduce various types of neural behavior. Because of their complexity, however, it is difficult to analyze the neural networks composed by these neuron models. Therefore, motivated by the desire to understand the efficient computational architecture underlying biological neural networks, spiking neural networks (SNNs), which are constructed with neuron models simplified for computation, have been researched. Among them, the leaky integrated firing (LIF) neuron model \cite{Gerstner2002} and its variants have been widely studied due to their mathematical tractability. SNNs are biologically plausible neural networks and differ significantly from general artificial neural networks (ANNs) in that they encode information into spikes and propagate the spikes through synapses. We focus on the dynamic temporal coding properties specific to neural circuits and SNNs in terms of the encoding scheme and the learning mechanism of the neural system. In the neural networks of biological organisms, it was shown experimentally that not only the spike firing rate but also the individual precise spike timing held information, and that it was an important factor in achieving more efficient computation \cite{Gollisch2008, Portelli2016}. In recent years, gradient methods for end-to-end supervised learning via direct spiking have been becoming established, but a number of research challenges remain due to the difficulties specific to SNNs.

One of the fundamental challenges in the supervised learning of SNNs using gradient methods is the non-differentiability of the discontinuous binary spike function. Various approaches have been developed and successfully overcome this difficulty \cite{Neftci2019, Nunes2022, Huh2018}. Most of those used in the recent research can be divided into two categories depending on a state quantity, through which the error signals are backpropagated: the rate-based, and the timing-based method. 
First, the rate-based method calculates the derivatives between time-evolving functions such as membrane potentials and spike functions, which leads to excellent results in a variety of tasks \cite{Bellec2020, Lee2020}. However, to achieve the functional derivatives, the rate-based method requires the time discretization of each function and the approximation of the gradient computation by replacing discontinuous spike gradients with continuous surrogate gradients (SG) during error backpropagation. In addition, there are concerns about how accurately the transfer of information is approximated due to errors introduced by these techniques of the rate-based method. Note that the use of SGs, which are continuous real-valued functions, can be interpreted as approximating the spike rate of each neuron with an activation function, suggesting that the network is not efficiently dealing with exact spike timing. Therefore, it is somewhat questionable whether it is appropriate as a temporal-coding scheme. 

In contrast, the timing-based method spearheaded by SpikeProp \cite{Bohte2000} provides a solution for computing exact gradients using spike timing. What is more, Mostafa \cite{Mostafa2018} explicitly derives spike timing gradients for non-leaky IF neuron models, leading to superior performance. Furthermore, Comsa et al. \cite{Comsa2020} and Göltz et al. \cite{Goltz2021} extended the method to LIF neuron models that also account for current leakage. The rate-based method backpropagates the gradient from the postsynaptic current to the presynaptic membrane potential regardless of the presence of spike firing events, whereas the timing-based method backpropagates the gradient between layers via the spike timing. Therefore, the timing-based method can avoid the problem of indifferentiability without the rate-based coding scheme. This point suggests the potential of the temporal-coding properties of timing-based methods. In general, timing-based methods adopt time-to-first-spike (TTFS) as a coding method, which converts the intensity of input information into absolute spike timing, and also they impose single-spike restrictions on all neurons \cite{Manea2022, Comsa2020, Goltz2021, Sakemi2021}. These ideas are supported by the physiological studies \cite{Rolls2006, Gollisch2008} that argue that the first spike retains more information than the later ones. On the other hand, we define \textit{multi-spike} SNNs as the networks in which individual neurons are allowed to fire multiple times, in contrast to SNNs with single-spike restrictions. It is clear that the single-spike restriction per neuron limits the learning efficiency and information processing capacity of SNNs. Furthermore, since biological neural networks are multi-spike, multi-spike is essential for investigating the coding properties of SNNs. Some researches tried on timing-based learning of multi-spike SNNs, but their results do not indicate that multi-spike SNNs were effective enough to outperform SNNs with single-spike restrictions \cite{Xu2013, Kim2020}. 

In response to the need to understand the characteristics of multi-spike SNN models, we propose a new algorithm for timing-based error backpropagation of multi-spike SNNs. To the best of our knowledge, this is the first study of supervised learning of SNNs based on spike timing allowing multi-spike. It is a natural extension of the timing-based method that originated from the Mostafa \cite{Mostafa2018} in that it adopted the LIF model as the neuron model and did not involve the intervention of heuristic techniques. 
The contributions of this paper are as follows:

\begin{itemize}
    \item We propose a timing-based backpropagation for multi-spike SNNs with exact gradients by constructing a learning algorithm via spike timing only.
    \item Using the MNIST dataset, a fundamental benchmark for image recognition, we revealed for the first time the fundamental properties of multi-spike dynamics, the relationship between the leakage time constant and the number of spikes, and its performance improvement over single-spike restricted timing-based methods.
    \item We propose a spike-count loss that can be used in gradient methods based solely on spike timing, called a \textit{dead neuron penalty} loss that pulls back neurons that do not respond to any input pattern into learning.
\end{itemize}

We hope that these results will provide a solid foundation for research on gradient learning methods for timing-based SNNs and serve as a good benchmark for working with more complex networks and tasks.

\section{Methods}

We propose a learning algorithm for feedforward SNNs based solely on spike timing. 
In this white paper, we employ the spike timing sequence as the output of each layer. Note that we use spike timing sequences based on two different representations of "spike order" depending on the situation: \textit{local spike order} and \textit{global spike order}. The former refers to the order in which spikes are output from a \textit{single} neuron. The local spike order-based spike timing of neuron $i$ is defined as $t_i^{(m)}$, which means the spike timing when neuron $i$ fires for the $m$-th time. The latter, on the other hand, is the spike order consistently counted for all output spikes from all neurons in a layer, which is an integration of the local spike order over all neurons in the layer. The spike timing based on the global spike order is defined as $\hat{t}^{(m)}$, which is the spike timing of the $m$-th spike in this layer. When the total number of spikes in a layer is $M$ and the number of output spikes for each neuron $i\in\{1, \cdots, I\}$ is $M_i$, these satisfy $M = \sum_{i=1}^{I} M_i$. 

\subsection{Neuron Models}

We use a leaky integrate-and-fire (LIF) neuron in continuous time. The neuron model involves reset dynamics, which is induced as an output spike is generated. Suppose that the local spike timing sequence of neuron $i\in\{1, \cdots, I\}$ in the previous layer is $\{t_i^{(m)}\}_{m=1}^{M_i}$. The differential equations satisfied by the postsynaptic current $I_j(t)$ and the membrane potential $V_j(t)$ of postsynaptic neuron $j\in\{1, \cdots, J\}$ at time $t\in \RR_{\geq0}$ are defined as
\begin{align}
    S_i(t) &= \sum_{m=1}^{M_i} \delta(t-t_i^{(m)}), \\
    \frac{\rd I_j(t)}{\rd t} &= -\frac{1}{\tau_I} I_j(t) + \beta_I \sum_{i=1}^{I} w_{ij} S_i(t), \\
    \frac{\rd V_j(t)}{\rd t} &= -\frac{1}{\tau_V} V_j(t) + \beta_V I_j(t), \\
    V_j(t)& \leftarrow 0 \quad ,\text{when}\ V_j(t) = V_{th}, 
\end{align}
where $S_i(t)$ is the spike sequence of the presynaptic neuron $i$, and $\delta(t)$ denotes Dirac delta function. $w_{ij}\in\RR^{I\times J}$ is a synaptic weight from presynaptic neuron $i$ to postsynaptic neuron $j$. $\tau_I\in\RR_{>0}$ and $\tau_V\in\RR_{>0}$ denote the decay time constants for the postsynaptic current and the membrane potential, respectively. $V_{th}$ is the threshold potential, which is fixed at $1.0$. $\beta_I\in\RR_{>0}$ and $\beta_V\in\RR_{>0}$ are the scale coefficients, but, for simplicity, we fix both $\beta$s to a $1$, hereafter. 

{\bf{Forward pass based on spike timing}}
In this study, a LIF model with decaying voltage and current with $\tau_I, \tau_V < \infty$ are adopted, and we let $\tau_I \neq \tau_V$ consistently. Assume here that neuron $j$ does not fire, i.e., this neuron satisfies the initial values $I_j(0), V_j(0) = 0$ at time $t=0$, continues to receive spikes from the previous layer. In this case, the postsynaptic potential of neuron $j$ at time $t$ and it's current, respectively, are obtained as convolution integral representations:
\begin{align}\label{eq:v_single}
    I_j(t) &= \int_0^{t} \theta(t') \mathcal{A}(t') \left(\sum_{i=1}^{I} w_{ij} S_i(t-t') \right) \rd t', \\
    V_j(t) &= \frac{\tau_{I} \tau_{V}}{\tau_{I}-\tau_{V}} \int_0^{t} \theta(t') \mathcal{K}(t') \left(\sum_{i=1}^{I} w_{ij} S_i(t-t')\right) \rd t', \\
    \theta(t) &= \left\{
    \begin{array}{ll}
    1 & \text{if} \quad t\geq0 \\
    0 & \text{otherwise}
    \end{array}
    \right. , 
\end{align}
where $\theta(t)$ is the step function and the kernel $\mathcal{K}$ and the function $\mathcal{A}, \mathcal{B}$ are defined as 
\begin{align}
    \mathcal{K}(t) &= \mathcal{A}(t) - \mathcal{B}(t), \\
    \mathcal{A}(t) &= \exp\left(-\frac{t}{\tau_{I}}\right), \\
    \mathcal{B}(t) &= \exp\left(-\frac{t}{\tau_{V}}\right). 
\end{align}
That is, the membrane potential $V_j(t)$ is represented as the weighted convolution integral of this kernel $\mathcal{K}$ with respect to all spikes in the output of the previous layer.

Then we consider the neuronal dynamics after the postsynaptic neurons fire. That is, it resets the membrane potential as an output spike is generated. Note that the synaptic current is not reset at this time. Unlike in the existing studies of SNNs using timing-based methods \cite{Mostafa2018, Comsa2020, Goltz2021}, where the single-spike restrictions let the activity of the neuron totally stopped at its own spike timing, the reset term results in multi-spike in this study.
Thereafter, $\hat{t}^{(m)}$ denotes the presynaptic spike timing at $m$-th in the global spike order, and the intersynaptic weight from the presynaptic neuron $i$ which fires at $m$-th in the global spike order to the postsynaptic neuron $j$ is defined as $w_{ij}^{(m)}$. Assuming that time $t$ satisfies $t\in[\hat{t}^{(m')}, \hat{t}^{(m'+1)})$ and that a neuron $j$ newly fires during this interval at $t = t_j^{(n'+1)}$. At any $t\in[\hat{t}^{(m')}, \hat{t}^{(m'+1)}) \land [t_j^{(n')}, t_j^{(n'+1)})$, the synaptic current and membrane potential of neuron $j$ can be written respectively as
\begin{align}\label{eq:v_sequence}
    I_j(t) &= \sum_{m=1}^{m'} w_{ij}^{(m)} \mathcal{A}(t-\hat{t}^{(m)}), \\
    \begin{split}
        V_j(t) &= \frac{\tau_{I} \tau_{V}}{\tau_{I}-\tau_{V}}
        \sum_{m=1}^{m'} w_{ij}^{(m)}  \left[\mathcal{A}(t-\hat{t}^{(m)})\right. \\
        &- \left.\max\left(\mathcal{A}(t_j^{(n')}-\hat{t}^{(m)}), \mathcal{B}(t_j^{(n')}-\hat{t}^{(m)})\right)\mathcal{B}(t-t_j^{(n')})\right]. 
    \end{split}
\end{align}

For convenience, we define the leaky factor of the two different time constants $\tau_{I, V}$ as $p = \tau_I / \tau_V$. Also, since spike timing only appears in an exponential form such as $\exp(-t/\tau_I), \exp(-t/\tau_V)$, we can transform times $t$ to $z$ according to the definition of $z := \exp(t/\tau_I)$, e.g., $\hat{z}^{(m)} := \exp(\hat{t}^{(m)}/\tau_I)$ and $z_j^{(n)} := \exp(t_j^{(n)}/\tau_I)$. Both $I_j, V_j$ at any $z\in[\hat{z}^{(m')}, \hat{z}^{(m'+1)}) \land [z_j^{(n')}, z_j^{(n'+1)})$ are multiplied by the spike timing sequence using $\{\hat{z}^{(m)}\}_{m=1}^{M}$ as
\begin{align}
    I_j(z) &= \sum_{m=1}^{m'} w_{ij}^{(m)} \frac{\hat{z}^{(m)}}{z}, \\
    \begin{split}
        V_j(z) &= \frac{\tau_{I} \tau_{V}}{\tau_{I}-\tau_{V}} \\
        &\sum_{m=1}^{m'} w_{ij}^{(m)} \left[\frac{\hat{z}^{(m)}}{z}-\frac{\hat{z}^{(m)} (\max\{z_j^{(n')}, \hat{z}^{(m)}\})^{p-1}}{(z)^p}\right], 
    \end{split} \\
    &= \frac{\tilde{\mathcal{A}}_j^{(m')}}{z} - \frac{\tilde{\mathcal{B}}_j^{(m', n')}}{(z)^p}, 
\end{align}
where, for convenience, we use the following definitions:
\begin{align}
    \tilde{\mathcal{A}}_j^{(m')} &= \frac{\tau_{I} \tau_{V}}{\tau_{I}-\tau_{V}} \sum_{m=1}^{m'} w_{ij}^{(m)} z^{(m)}, \\
    \tilde{\mathcal{B}}_j^{(m', n')} &= \frac{\tau_{I} \tau_{V}}{\tau_{I}-\tau_{V}} \sum_{m=1}^{m'} w_{ij}^{(m)} \hat{z}^{(m)} (\max\{z_j^{(n')}, \hat{z}^{(m)}\})^{p-1}. 
\end{align}
For each case, a solution for the spike time $z_j^{(n'+1)}$, defined by
\begin{align}
    V_j(z) = V_{th}
\end{align}
has to be found, sequentially. In the whole experiments in the study, let the leaky factor $p = 2$. Therefore, we can solve the next spike timing $z_j^{(n'+1)}$ can be solved explicitly and analytically by using the coefficient parameters $\tilde{\mathcal{A}}_j^{(m')}, \tilde{\mathcal{B}}_j^{(m', n')}$ based on the historical spike timing sequence of the previous layer $\{z^{(m)}\}_{m=1}^{m'}$ and its own previous spike timing $\hat{z}_j^{(n')}$ as 
\begin{align}
    \label{eq:update}
    z_j^{(n'+1)} &= \frac{\tilde{\mathcal{A}}_j^{(m')} - \sqrt{\left(\tilde{\mathcal{A}}_j^{(m')}\right)^2 - 4V_{th} \tilde{\mathcal{B}}_j^{(m', n')}}}{2 V_{th}}. 
\end{align}
The condition for the existence of $z_j^{(n'+1)}$ is that each coefficient parameter $\tilde{\mathcal{A}}_j^{(m')}, \tilde{\mathcal{B}}_j^{(m', n')}$ and the root content of the equation (\ref{eq:update}) are non-negative and $z_j^{(n'+1)} \in [z^{(m')}, z^{(m'+1)})$. When actually implementing the algorithm, it can be computed independently for each $m',  j$ in accordance with the definition of $z_j^{(n'+1)}$. The obtained spike timing sequences $\{z_j^{(n')}\}_{n'}$ for each neuron are finally merged into a unified spike timing sequence $\hat{z}^{(n)}$ in the global spike order, and then subsequent layers can be calculated in the same way.

\begin{figure*}
    \centering
    \includegraphics[width=.9\linewidth]{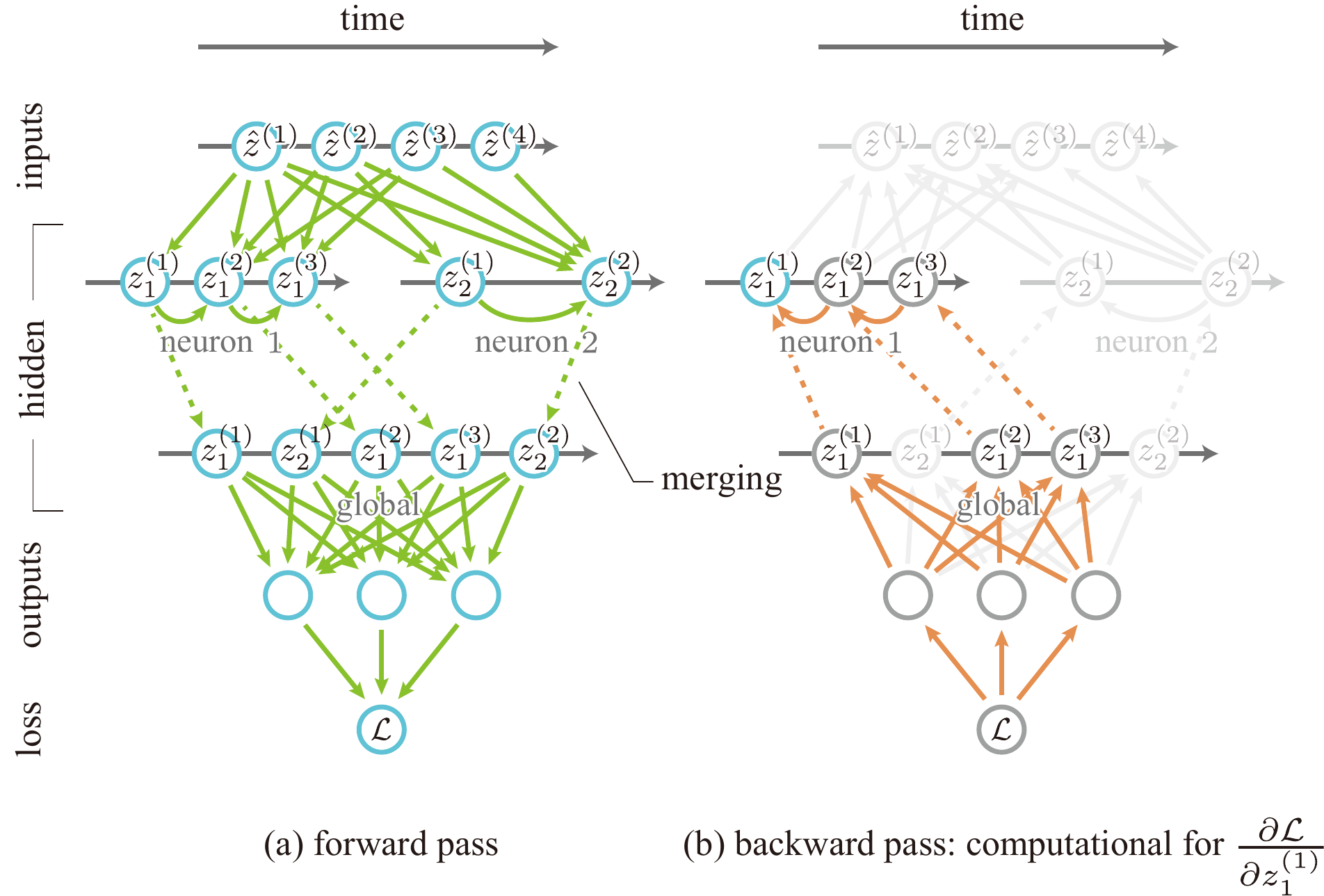}
    \caption{Computational graph of a multi-spiking neural network (4-2-3). Inputs and hidden units are spike units. The SNN has only two neurons in the hidden layer and, in this case, neuron $1$ and neuron $2$ fire three times and twice in a given period, respectively. (a) The forward pass is illustrated as colored arrows: the green solid arrows indicate the propagation of spikes through the synapse or the reset path, the green dashed arrows indicate merging the spikes of each neuron in the global spike order. (b) Backpropagation flow of computational for $\frac{\partial \mathcal{L}}{\partial z_1^{(1)}}$ with orange arrows. }
    \label{fig:snn_flow}
\end{figure*}
Fig. \ref{fig:snn_flow} shows the computational graph where edges start from the input spike sequence to the loss function and its backward. In the hidden layer, the computation of the spike timing is done independently for each neuron and finally merged into a spike timing sequence based on the global spike order for the entire layer (the dashed arrows). The arrows between spikes within the same neuron represent the reset paths derived from their own most recent past spike timing, which are induced by the reset of the membrane potential. When calculating the hidden layer spike timing sequence from the input spike timing sequence, the computational path from the input spike received earlier will continue to affect the calculation of spike timing after that received time via two types of the path: the postsynaptic currents that remain after resets, and the latest spike timing of themselves. Note that, considering the convolution integral representations above, the time range that a neuron is affected by a previous single spike depends on the leakage time constant. In the case of a multi-layered SNNs, the spike timing sequence based on the global spike order of the previous hidden layer can be regarded as the input spike sequence, and the spike timing sequence of the next layer can be calculated just as the spike timing is calculated from the input spike sequence in the figure. The second graph illustrates to computation for $\frac{\partial \mathcal{L}}{\partial z_1^{(1)}}$ by backpropagation from the bottom 
 $\mathcal{L}$ (the orange arrows).  Note that the gradient is propagated backward without ignoring all the reset paths.

\subsection{Input formats}

The input format to the network is latency coding. The latency input consists of the same number of input neurons as the dimension of the input data, and each value is mapped to the spike timing of each input neuron in a one-to-one fashion. Each input neuron fires only once per sample, and its exact spike timing contains the information. Every input neuron has a one-way weighted, fully connected, directed path to all neurons in the hidden layer, and spikes carrying information at the spike timing are input as synaptic currents through this path to each middle layer.

\subsection{Loss functions}

We adopt membrane potential loss as the loss function. The same number of output neurons as a given number of output dimensions is placed in the final layer, and the output from one neuron corresponds to one output dimension. The threshold is set to infinite only for the output neurons, and they never fire. The value of the membrane potential of each output neuron was observed at the preset end time $t_{out}$, and this value was taken as the final output of each neuron. In other words, the output membrane potential $\textbf{v}_{out}:=\{v_o\}_{o=1}^{N}$ of this network is defined by the output layer coefficient parameters $\mathcal{A}_{o}^{(M)}, \mathcal{B}_{o}^{(M, 0)}$ as
\begin{align}
    v_o = \frac{\mathcal{A}_{o}^{(M)}}{z_{out}} - \frac{\mathcal{B}_{o}^{(M, 0)}}{(z_{out})^2}, 
\end{align}
where $z_{out} := \exp(t_{out}/\tau_I)$. For the multiclass classification problem, the cross-entropy, which incorporates the correct answer labels into the softmax of these output values, was used as the error function.

In this paper, we also introduce \textit{spike count penalty} loss for efficient learning. By incorporating this into the overall model loss function, the number of spikes in each neuron can be freely manipulated even in the gradient method for SNNs based only on spike timing. This penalty loss provides one solution to the challenge that one of the difficulties in learning timing-based SNNs is that it is unable to incorporate the spike counts directly into the model because the timing-based error backpropagation has no state variables other than the spike timing.

To incorporate the spike count penalty loss, the number of spikes for each neuron that fires during the forward trial process is counted, and a flag
\begin{align}
    \mathcal{P} := \left\{\mathcal{P}\in\{0,\ 1\}^{J}\ |\ \mathcal{P}_j = c(M_j),\ \forall j=1, \dots, J \right\}
\end{align}
is the vector for each neuron that meets the specified spike count condition, where $M_j$ is the spike count of neuron $j$ and $J$ is the number of neurons in the target hidden layer. $c: \ZZ_{\geq0} \rightarrow \{0, 1\}$ is a binary function that defines the spike count condition. Let $\textbf{v}_{hidden}$  be a $J$-dimensional vector of the measured values of each output neuron membrane potential at the end time of the trial $t_{out}$, and then the spike count penalty loss, 
\begin{align}
    L_{c}(\textbf{v}_{hidden}; \mathcal{P}) = \frac{1}{J}\mathcal{P}^{\top}(-\textbf{v}_{hidden}+V_{th} \textbf{1}_{J})
\end{align}
is obtained, where $\textbf{1}_{J}$ is an $J$-dimensional 1-vector. By incorporating this penalty loss, the spike counts of individual neurons can be conditioned into the SNN timing-based learning algorithm. 

In particular, by penalizing \textit{dead neurons} that do not fire for any input pattern, it is possible to suppress initial value dependence, which has been considered difficult in timing-based methods, and to facilitate the experimental design. In this case, we refer to the loss of spike count penalty as 'dead neuron loss'. Importantly, by setting the conditional equation to flag dead neurons, we can avoid the problem of neurons that do not fire depending on the initial weight variable (since the timing-based method only handles spike timing, neurons that never fire cannot be included in the training, and all gradients will be zero, making them untrainable). This reduces the dependence on the initial weight variables, which is a major concern with timing-based methods, and makes SNN training design much easier.

The square of the norm of the output multiplied by a small factor is also added to the overall loss to prevent output values from diverging. Although learning is possible without this loss term, we added it in our experiments because we observed that the loss function of the test increased in the middle of learning, even though the accuracy increased.

In summary, for the obtained network membrane potential output $v_o$ and target class indicator probability $q$, the error function $E$ is
\begin{align}
    E(\textbf{v}_{out}) &= - \sum_{o} q_o \log \text{softmax}(v_o), 
\end{align}
and the loss function $\mathfrak{L}(\textbf{v}_{out})$ for the whole model is given by using the hyperparameters $\lambda,\ \sigma$ as
\begin{align}
    \mathfrak{L}(\textbf{v}_{out}) &= E(\textbf{v}_{out}) + \lambda L_{c}(\textbf{v}_{hidden}; \mathcal{P}) + \sigma \frac{1}{N_{out}} ||\textbf{v}_{out}||_2^2. 
\end{align}

\subsection{Memory- and computation-efficient algorithms}

Training and inference of SNNs on ordinary Von Neumann type computers are extremely computation- and memory-intensive due to the nature of the dynamics in continuous time. Although research on SNNs has been accelerated in recent years with the rapid development of computers, its computational complexity still remains an obstacle to research on SNNs. In fact, this limitation has made it difficult to handle large-scale datasets and models. To realize a computationally efficient timing-based error backpropagation method that allows multi-spike, we adopted a computation cycle based on the spike count independent of the spike timing and introduced the two-stage partitioned algorithm. First, the spike-count-based computation loop enables parallel processing of calculations proportional to the input dimension and the number of hidden neurons, which can be efficiently computed on GPUs and TPUs to achieve a fast computation rate. Furthermore, though there is a trade-off between computation time and memory resources, and if the input dimension or model parameters are large, the burden on memory increases when covering a large number of spikes, the two-stage cycle algorithm can dramatically reduce the amount of memory required for computation. The sequence of processes of the proposed algorithm with two-stage cycle processing is shown in Algorithm 1. In this algorithm, the iterator $j, m'$ loops can be computed in parallel.
\begin{figure}[!t]\label{alg:1}
    \begin{algorithm}[H]
        \caption{Forward pass of multi-spiking neural networks}
        \begin{algorithmic}[1]
        \REQUIRE $\{\hat{t}^{(m)}\}_{m}$: Vector of input spike timing
        \REQUIRE $M$: Number of input spikes
        \REQUIRE $J$: Number of hidden neurons
        \REQUIRE $O$: Number of output neurons
        \REQUIRE $\{w_{i j}\}_{i, j}, \{w_{j o}\}_{j, o}$: Set of weight matrix
        \REQUIRE $N_1,\ N_2$: Maximum number of spikes to break the loop
        \REQUIRE $t_{out}$: Time to observe output membrane potential
        \REQUIRE $\tau_I$: Leakage time constant
        \ENSURE  $\{v_{o}\}_{o}$: Vector of output membrane potential
        \\ \textit{Initialisation} :
            \STATE $\hat{z}^{(m)} \leftarrow \exp(\hat{t}^{(m)}/\tau_I),\ \forall m\in\{1, \cdots, M\}$
            \STATE $z_j^{(0)}[m'] \leftarrow 0,\ \forall j\in\{1, \cdots, J\},\forall m'\in\{1, \cdots, M\}$
            \STATE $z_{out} \leftarrow \exp(t_{out}/\tau_I)$
        \\ \textit{1st Loop Process} : 
            \FOR {$n' = 0$ to $N_1-1$} 
            \FOR {$j = 1$ to $J$ \textbf{(Parallel)}}
            \FOR {$m' = 1$ to $M$ \textbf{(Parallel)}}
            \STATE $\tilde{\mathcal{A}}_j^{(m')} \leftarrow \tau_I \sum_{m=1}^{m'} w_{ij}^{(m)} z^{(m)}$
            \STATE $\tilde{\mathcal{B}}_j^{(m', n')} \leftarrow  \tau_I \sum_{m=1}^{m'} w_{ij}^{(m)} \hat{z}^{(m)} \max\{z_j^{(n')}, \hat{z}^{(m)}\}$
            \IF {$\tilde{\mathcal{B}}_j^{(m', n')}, \tilde{\mathcal{A}}_j^{(m')}, \left(\tilde{\mathcal{A}}_j^{(m')}\right)^2 - 4 \tilde{\mathcal{B}}_j^{(m', n')} V_{th} \geq 0$}
            \STATE $z_j^{(n'+1)}[m'] \leftarrow \frac{\tilde{\mathcal{A}}_j^{(m')} - \sqrt{\left(\tilde{\mathcal{A}}_j^{(m')}\right)^2 - 4 \tilde{\mathcal{B}}_j^{(m', n')} V_{th}}}{2V_{th}}$
            \IF {$z_j^{(n'+1)}[m'] < \hat{z}^{(m')}$ \OR $z_j^{(n'+1)}[m'] > \hat{z}^{(m'+1)}$}
            \STATE $z_j^{(n'+1)}[m'] \leftarrow \infty$
            \ENDIF
            \ELSE
            \STATE $z_j^{(n'+1)}[m'] \leftarrow \infty$
            \ENDIF
            \ENDFOR
            \STATE $z_j^{(n'+1)} \leftarrow \min_{m'}z_j^{(n'+1)}[m']$
            \STATE $\mathcal{A}_{o} \leftarrow \mathcal{A}_{o} + \beta_I \beta_V \tau_I\ w_{jo} \left(z_j^{(n'+1)}\right)^2,\ \forall o\in\{1, \cdots, O\}$
            \STATE $\mathcal{B}_{o} \leftarrow \mathcal{B}_{o} + \beta_I \beta_V \tau_I\ w_{jo}\ z_j^{(n'+1)},\ \forall o\in\{1, \cdots, O\}$
            \ENDFOR
            \ENDFOR
            \\ \textit{2nd Loop Process} : 
            \FOR {$n' = N_1$ to $N_2-1$} 
            \FOR {$j \in \{j\ |\ z_j^{(N_1)} < \infty\}$ \textbf{(Parallel)}} 
            \STATE the same process as 1st Loop 
            \ENDFOR
            \ENDFOR
        \STATE $v_{o} \leftarrow \frac{\mathcal{A}_{o}}{z_{out}} - \frac{\mathcal{B}_{o}}{(z_{out})^2},\ \forall o$ 
        \RETURN $\{v_{o}\}_{o}$ 
        \end{algorithmic} 
    \end{algorithm}
\end{figure}

\section{Results}

We use the MNIST benchmark to illustrate the high performance of our method and the properties of SNNs acquired by multi-spike. Our training model is characterized by the application of error backpropagation to a feed-forward SNN with multi-spike, without using surrogate gradient or plasticity. The weight variables were initialized according to a Gaussian distribution with expectation and variance of $0.03$ and $0.3$, respectively, for all tasks. There were $10$ output units since all experiments were solving MNIST classification problems, and the inputs were composed of the same number of input neurons and spikes as image dimensionality $784(=28\times28)$. The number of neurons in the hidden layer is consistently set to $400$, making a two-layer network ($784-400-10$ network). Weights were updated using Adam \cite{Kingma2015} and trained $100$ epochs with a learning rate of $0.001$. With respect to the input, the MNIST dataset consists of $70000$ grayscale handwritten digit images and correct answer labels consisting of any integer from $0-9$. Of these, $60000$ were used for training and the remaining $10000$ for testing. The pixel intensity in each dimension was scaled to a value between $[0,1]$ after inverting the black and white of the image and converted to input spike timing between $t=0.0$ and $t=1.0$. This was split into mini-batches of $100$ each and trained. The output was the membrane potential of the output neuron at the end time $t_{\text{end}} = 1.0$. As a dead neuron loss, we flagged neurons that only fire on input patterns less than $1/10$ of the mini-batch size. Hyperparameters were set as $\lambda = 0.01, \sigma = 0.0001$ after grid-searched, but their effect on learning performance was small unless taken to extremes. Bias spiking was not used in any of the experiments. All programs were implemented on PyTorch \cite{Paszke2019}.

\begin{figure*}
    \centering
    \includegraphics[width=.9\linewidth]{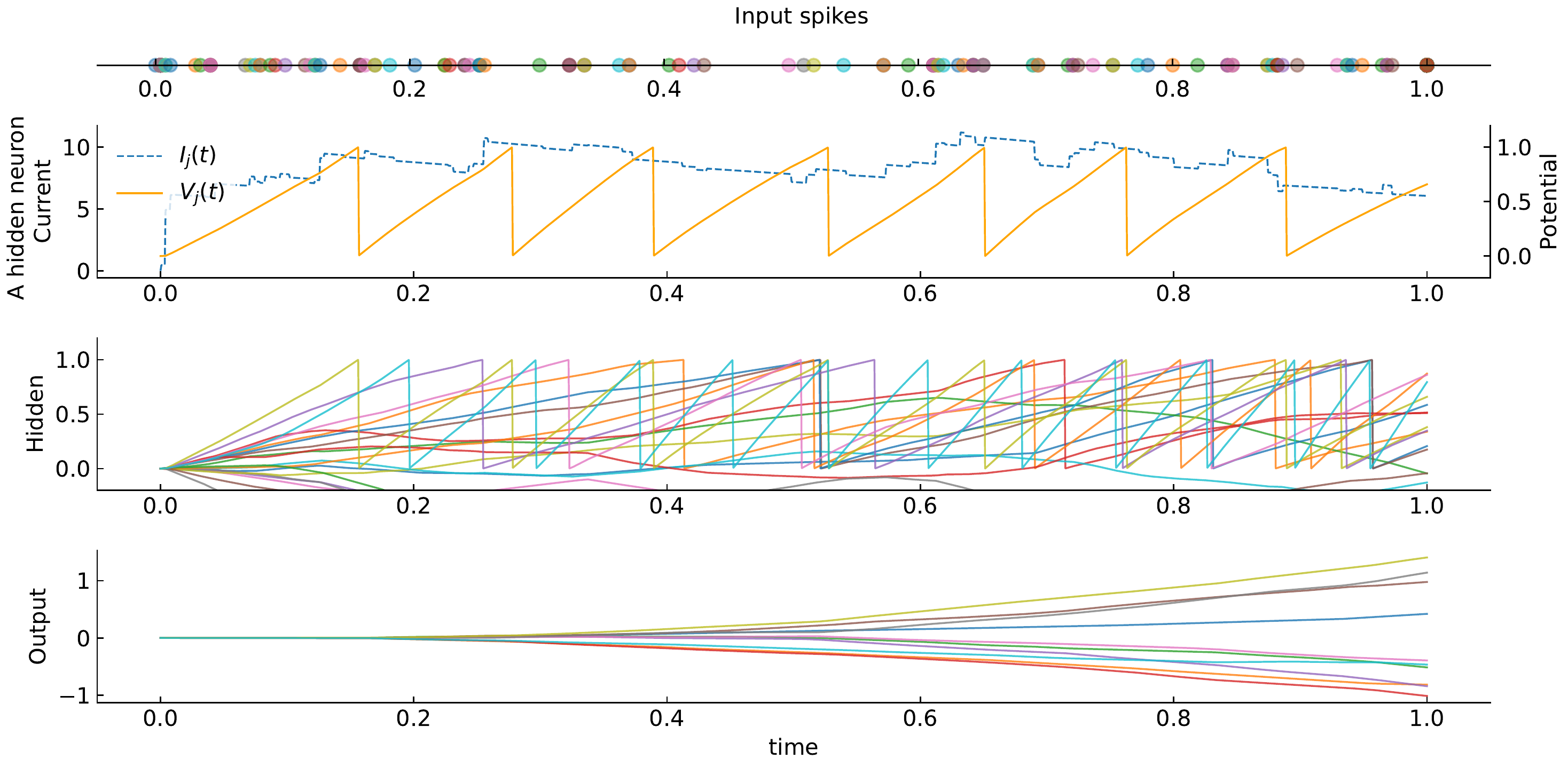}
    \caption{Time evolution of the membrane potential and postsynaptic current of each neuron when a certain input pattern is given to the learned network. At the top, a single round symbol indicates a single input spike. The second from the top shows the membrane potential and postsynaptic current of a neuron selected in the hidden layer, represented by solid and dotted lines, respectively. The third figure from the top captures the temporal evolution of the membrane potential of 20 neurons in the hidden layer. The bottom figure illustrates the membrane potential of each of the 10 output neurons over time.}
    \label{fig:potential_time}
\end{figure*}
Fig. \ref{fig:potential_time} shows the dynamics of neurons in the hidden layer and the output layer when an input spike sequence was input to a trained SNN with $\tau_{I} = 1.0$ under the above parameter settings. The number of spikes for each neuron varies significantly, and it can be seen that the firing frequency of the same neuron fluctuates along with the postsynaptic current within a given time ($t_{out} = 1.0$ in this case).

First, we investigated how the number of spikes of SNNs can be altered under multi-spike conditions. In the two-layer feedforward SNNs described above, the entire computational process is based solely on the spike timing of each spike. It is worth noting what distribution of the number of spikes per neuron emerges when multi-spike is allowed.
\begin{figure}
    \centering
    \includegraphics[width=.9\linewidth]{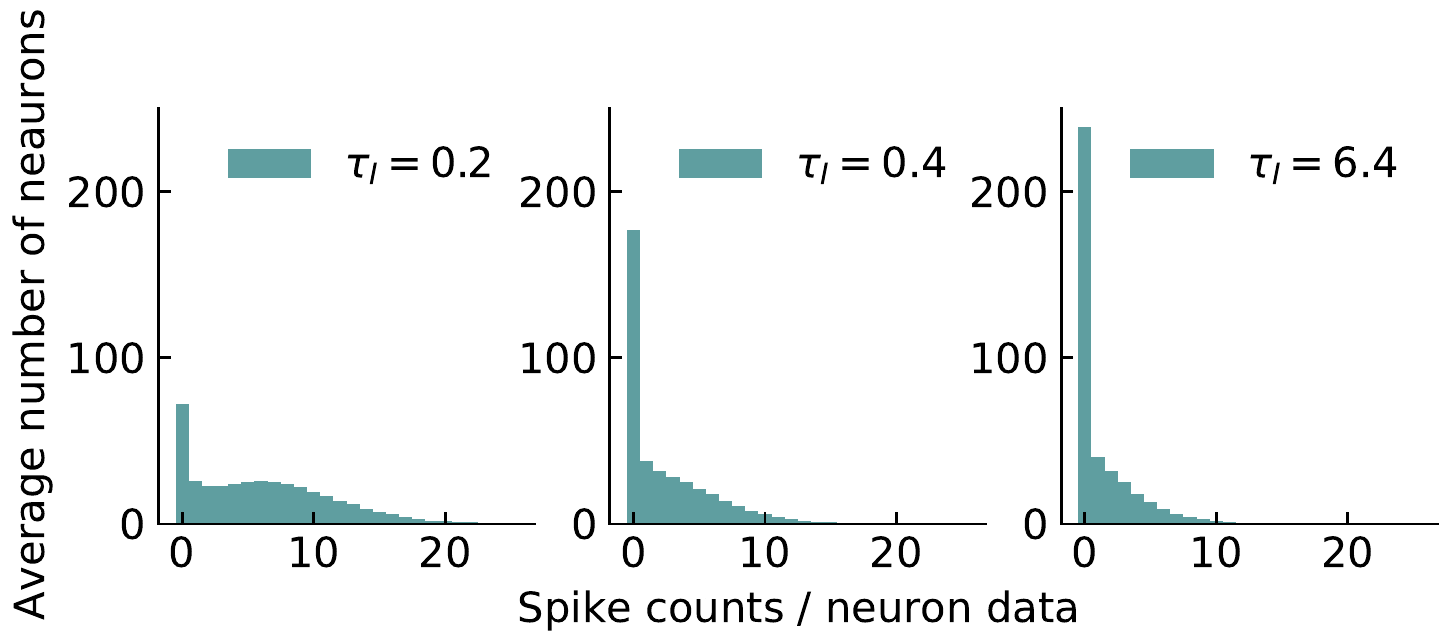}
    \caption{Distributions of the number of spikes in SNNs (784-400-10) for various leakage time constants $\tau_{V} = 0.2,0.4,6.4$. }
    \label{fig:spikenum_dist}
\end{figure}
Fig. \ref{fig:spikenum_dist} shows histograms of the number of spikes of each hidden neuron, which is averaged over ten different SNNs trained with different initial weights. As this figure shows, the distribution depends on the magnitude of the current and voltage leakage terms, i.e., the leakage time constant. Fig.\ref{fig:spikenum_dist} shows that the smaller $\tau_I$ is, the larger the mode of the distribution of the number of spikes shifts toward. For $\tau_I = 0.2$, the distribution shifts unimodal to multimodal and the higher mode of the distribution appears separated into non-$0$ areas. This indicates a bifurcation of all neurons in an SNN into two unconsolidated classes, neurons that do not fire at all and neurons that do fire at least once, which significantly hinders learning. 

\begin{figure}
\centering
\includegraphics[width=.9\linewidth]{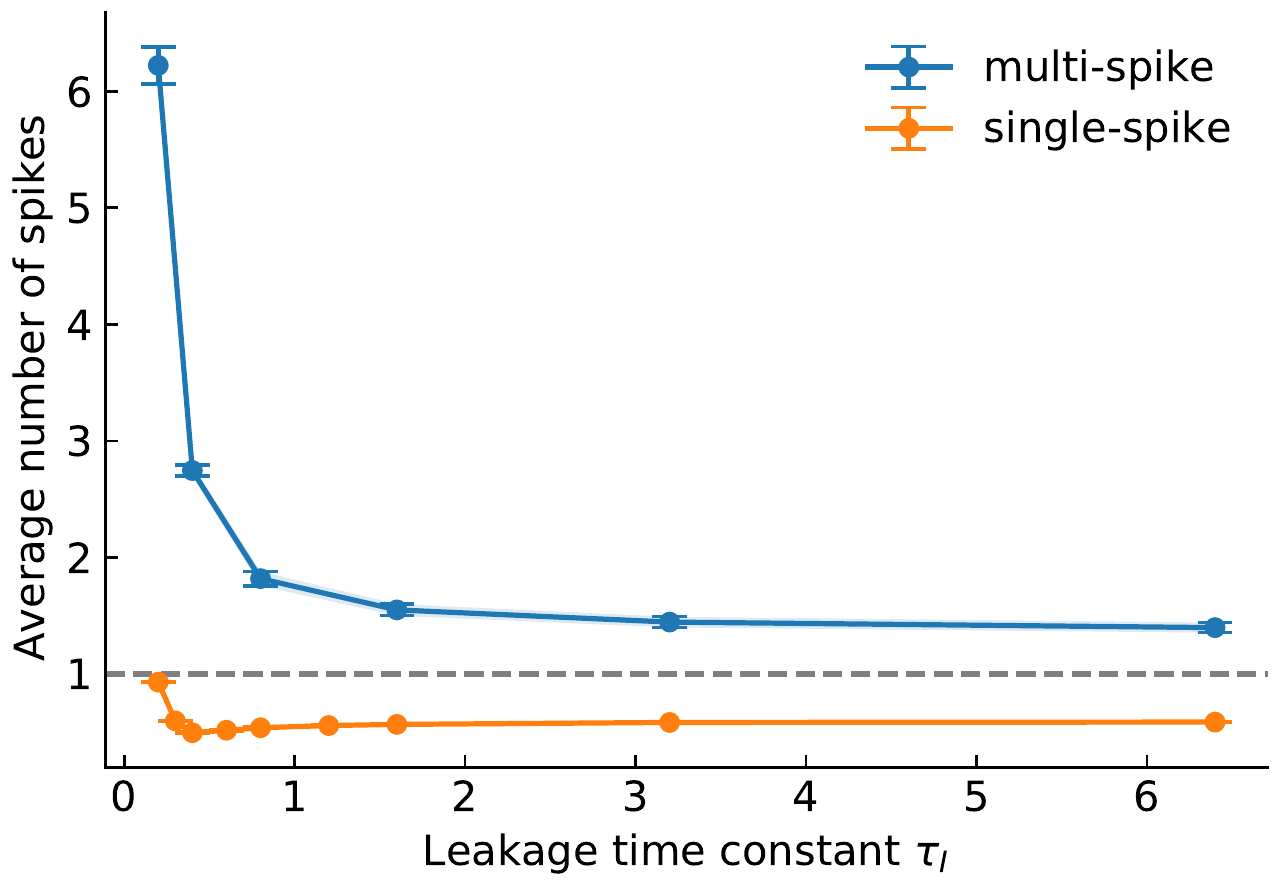}
\caption{The relationship between the average number of spikes per neuron and the leakage constant time $\tau_I$. the dashed horizon line denotes $1$, which the SNN model with single-spike restrictions cannot exceed. }
\label{fig:spikenum-tau}
\end{figure}
Fig. \ref{fig:spikenum-tau} shows the spike count per hidden neuron in each trained SNN when the leakage time constant is varied, averaged over all hidden neurons. Fig. \ref{fig:spikenum-tau} indicates that the smaller the time constant of the synaptic leakage term, i.e., the larger the leakage rate, under conditions in which multi-spike are allowed, the more average number of neurons fire. By contrast, as the leakage time constant increases, i.e., as the model asymptotically approaches a non-leaky neuron model, the average number of spikes per neuron converges to a certain value $\simeq 1.5$. With the single-spike restrictions, the average number of spikes per neuron is equal to the density of the entire network, the proportion of neurons that fire with respect to an input pattern. The figure with single-spike restrictions indicates that when the leakage time constant approaches close to zero, almost all neurons will fire. The SNNs with the single-spike restrictions converge to 0.6 asymptotically (non-leaky models), but unlike the multi-spike model, the curve is neither smooth nor monotonic. In terms of sparsity on a per-neuron basis, the multi-spike SNNs are sparser than the SNNs with single-spike restrictions over a wide range of leakage time constants.

\begin{figure}
    \centering
    \includegraphics[width=.9\linewidth]{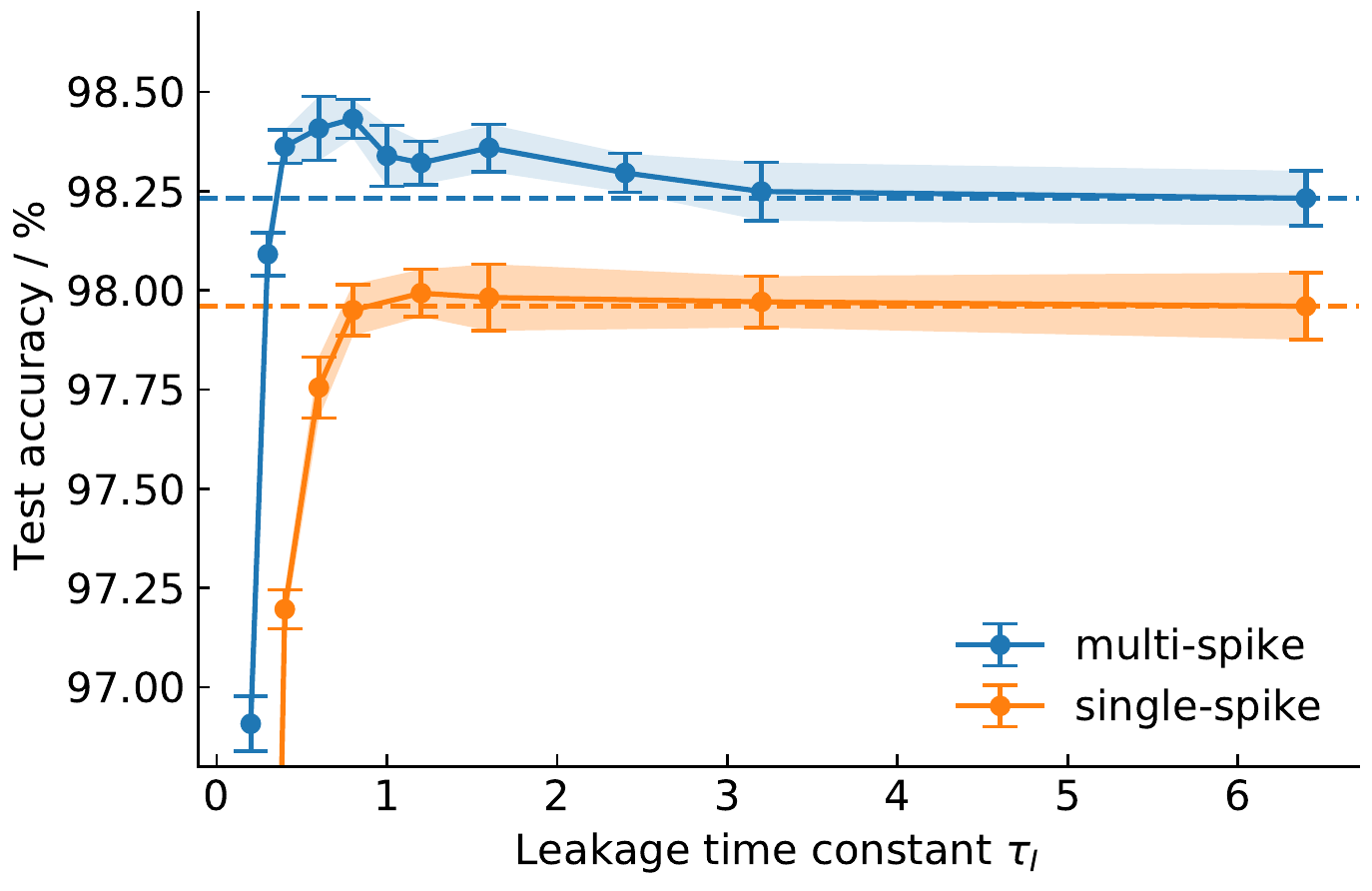}
    \caption{The relationship between learning performance and the leakage constant time $\tau_I$.}
    \label{fig:accuracy-tau}
\end{figure}
Fig. \ref{fig:accuracy-tau} shows the relationship between leakage time constant and test accuracy on the MNIST dataset. First, the multi-spike SNNs outperform the SNNs restricted to single-spike in accuracy under any leakage time constant condition. Common to both multi-spike and single-spike SNNs is that as the leakage time constant is very close to zero accuracies deteriorate rapidly as leakage increases. For both cases, there is a convergence value when the leakage time constant is increased, converging to $98.23 \%$ for multi-spike and $97.96 \%$ for single-spike, respectively. Note that there is a clear optimum leakage time constant $\tau_I=0.8$ (with the test accuracy at $98.43\pm0.05\%$) only for the multi-spike case. 

\begin{table*}[t]
    \caption{Performance of supervised learning for neural networks on MNIST tasks by neuron type}
    \label{tbl:comp_acc}
    \centering
    \begin{tabular}{cccccccc} \hline
        Network & Coding & BP scheme & Spike type & Leakage & $\tau_V$ & $\tau_I$ & Test accuracy \\ \hline
        784-800-10 (+ dropout)\cite{Srivastava2014} & ANN & & & & & & $98.4\%$ \\ \hline
        784-800-10\cite{Lee2016} & SNN (rate coding) & surrogate-gradient & multi & leaky & $\tau$ & $\tau$ & $98.64\%$ \\
        784-800-10\cite{Jin2018} & SNN (rate coding) & surrogate-gradient & multi & leaky & $\tau$ & $\tau$ & $98.84\pm0.02\%$ \\
        \hline
        784-800-10 \cite{Mostafa2018} & SNN (temporal coding) & timing-based & single & non-leaky & $\infty$ & $\tau$ & $97.55\%$ \\
        784-340-10 \cite{Comsa2020} & SNN (temporal coding) & timing-based & single & leaky & $\tau$ & $\tau$ & $97.96\%$ \\ 
        784-500-10 \cite{Sakemi2021} & SNN (temporal coding) & timing-based & single & non-leaky & $\infty$ & $\infty$ & $97.99\pm0.07\%$ \\ 
        784-350-10 \cite{Goltz2021} & SNN (temporal coding) & timing-based & single & leaky & $\tau$ & $2\tau$ & $97.2\pm0.1\%$ \\ 
        256-246-10 \cite{Cramer2022} & SNN (temporal coding) & surrogate-gradient & multi & leaky & $\tau$ & $\tau$ & $97.5\pm0.1\%$ \\
        784-400-10 [This work] & SNN (temporal coding) & timing-based & single & non-leaky* & $\tau$ & $2\tau$ & $97.96\pm0.08\%$ \\ 
        784-400-10 [This work] & SNN (temporal coding) & timing-based & single & leaky & $\tau$ & $2\tau$ & $\bf{97.99\pm0.06\%}$ \\ 
        784-400-10 [This work] & SNN (temporal coding) & timing-based & multi & non-leaky* & $\tau$ & $2\tau$ & $98.23\pm0.07\%$ \\ 
        784-400-10 [This work] & SNN (temporal coding) & timing-based & multi & leaky & $\tau$ & $2\tau$ & $\bf{98.43\pm0.05\%}$ \\ \hline
    \end{tabular}
    
    *It can be regarded as non-leaky due to extreme conditions: $\tau = 3.2$.
\end{table*}
A comparison of the results obtained in this experiment with the MNIST test accuracy of various existing 2-layer feedforward SNNs are summarized in Table \ref{tbl:comp_acc}. The error indications in our study are defined by the standard errors of the results of 10 experiments performed with different initial weights. The results show that in supervised learning of simple feedforward SNNs trained with temporal coding, multi-spike SNNs in the timing-based method achieve the highest accuracy over all existing methods. In addition, there is already no significant difference when compared to SNN supervised learning based on rate coding.

\section{Discussions}

We performed supervised learning on both the existing single-spike-limited SNNs per neuron and the multi-spike-allowed model in this study and compared the properties of the error backpropagation method for SNNs based solely on strict spike timing. In particular, the present study shows that omitting the single-spike restriction and allowing multiple firings of a neuron improves the performance of the standard MNIST benchmark under all leakage time constant conditions. A notable feature of our method is that it uses the exact spike timing in the backpropagation computation process, thus achieving ideal temporal coding mediated by spikes on multi-spike SNNs in terms of gradient computation. In related works on multi-spike SNNs, although types of SG method, applied an appropriate regularization and achieved sparse, near-temporal-coding learning \cite{Cramer2022, Yan2022}. However, since these learning methods using gradient approximation assume continuous spiking behaviors during backpropagation, it is open to question whether they are strictly temporal-coding learning from the viewpoint of gradient propagation. To the best of our knowledge, our proposed algorithm, for the first time, achieved the supervised learning of multi-spike SNNs with backpropagation of exact gradients based on spike timing.

Our proposed model has advantages in precisely dealing with the reset dynamics. The resetting behavior at the moment of firing is one of the factors that significantly influence the information processing mechanism. In general, studies addressing supervised learning of multi-spike SNNs have introduced double-reset neuron models, which reset the membrane potential and also the corresponding postsynaptic current at the same time due to its simplicity \cite{Bellec2020, Kim2020}. However, there are concerns that the adoption of double-reset means that any information input up to each spike timing is lost on each firing, completely breaking the dependency between previous and subsequent spikes within the same neuron, thus lacking the benefits of multi-spike. By contrast, we reset only the membrane potential that has reached the threshold to the baseline, $0$, while the postsynaptic current is not reset and retains its pre-firing values. In this way, the past information received by the neuron itself is propagated into the future via both its own past spike timing and the postsynaptic current path. There is a non-negligible difference in the network structure whether double-reset is employed or not, and we employed the latter in this study.

Heuristic approximations and abbreviations in the error backpropagation calculation process for SNNs are also major factors that compose the network learning paradigm. In general, learning methods using surrogate gradients ignore the error backpropagation paths led by resets of membrane potential, what is called reset path, in multi-spike \cite{Zenke2021, Bellec2020, Kim2020}. Backpropagating through surrogate gradient results in the fact that the reset information is propagated recursively by a number of times equal to the number of time steps, even though the actual reset is instantaneous at the spike timing. Even though the amount of the approximation error induced by a single surrogate derivative is tolerable, the accumulated error can be huge because the number of time steps is usually by far larger than the number of spikes \cite{Zenke2021, Kim2020}. To avoid this approximation error diverging is the main reason why they ignore the reset paths. On the other hand, the timing-based method used in this study does not use any approximation in the whole calculation process of backpropagation, i.e., it realizes an exact gradient calculation. It should be noted, however, that the timing-based method cannot capture the occurrence and disappearance of sudden spikes caused by weight updates and the accompanying changes in the number of spikes. The sudden spike changes pointed out here are not spike changes that increase or decrease with continuous changes in membrane potential in the time direction around the end time $t_{out}$, but spike changes that occur with discontinuous membrane potential changes in the middle of a given period $(0, t_{out})$. Although this is a challenge inherent to the timing-based method as pointed out by SpikeProp \cite{Bohte2000}, it was experimentally shown that learning works well both for single-spike SNNs and also for the multi-spike SNNs in this study. The impact of this sudden spike variation is one of the issues that should be considered in the future.

Through this study, it is clear that the properties of the overall network spike count are altered depending on the time constants of the membrane potential and synaptic current, i.e., the synaptic leakage term. The results basically show that the smaller the leakage time constant (i.e., the more leakage), the higher the overall network spike count, which leads to higher learning performance. This change in performance with increasing voltage leakage is a phenomenon not seen in existing SNNs restricted to single-spike. In the case of multi-spike SNNs trained through surrogate gradients, through experiments of rate-coding with feedforward networks, \cite{Chowdhury2021} showed that the leakage of membrane potentials and postsynaptic currents ensures the robustness of the network, leading to high learning performance. In the present timing-based method, a similar result may be derived from the correlation between the increase in the number of spikes and the increase in performance. In other words, it is possible that the increased rate-coding-like quality induced by the increase in the number of spikes may contribute to the superior generalization effect. In contrast, single-spike SNNs are clipped at a maximum of $1$ in the number of spikes per neuron, so the amount of leakage of the membrane potential should not affect their performance. Considering that time constants smaller than the time scale of the experiment ($1 \text{ms}$ in this case) would conversely cause a sharp decline in performance, therefore, there are optimal values for the time constants of membrane potential and synaptic current for multi-spike SNNs. In particular, the amount of optimized leakage time constant can be characterized by the magnitude of the task itself imposed on the SNNs and the local network structure on which the time scale of the experiment depends. Optimizing the learning capacity of the entire network through leakage cannot be discussed only in terms of machine learning aspects. It has been studied and reported from a physiological perspective that the leakage time constant of a neuron plays an important role as an indicator to characterize that brain region and that the diversity of time constants is key to the task-solving ability of that neural networks \cite{Kim2021, Manea2022}. This is in line with the relationship derived in this study between the optimal leakage time constant and the timescale of the tasks.

\section{Conclusion}

In this study, we proposed a novel efficient backpropagation algorithm for multi-spike SNN models and pursued the goal of achieving high algorithmic performance and clarifying its learning characteristics compared to single-spike SNN models. Moreover, we proposed novel learning techniques such as a two-stage algorithm based on spike count and the addition of a dead neuron penalty. The performance of our algorithm on the MNIST dataset outperformed the state-of-the-art SNNs based on temporal coding and was confirmed to be on par with the conventional SNNs and ANNs based on rate coding. Focusing on the leakage time constant of the postsynaptic current and the membrane potential, we investigated how this affected the characteristics of spike counts of SNNs with multi-spike neurons. We also showed the existence of an optimal membrane potential leakage rate that achieved maximum performance, around the time scale of the task. Moreover, the phenomenon was not observed in training SNNs with single-spike restrictions. Our multi-spike networks suggested the possibility of supervised learning based on the spike timing of SNNs with a biologically plausible recurrent topological structure. In the future, we would like to pursue timing-based learning of more complex and large-scale tasks and recurrent networks. Finally, we would like to add that our SNN model, which learns with information encoded in the spike timing, can also contribute to the construction of very energy-efficient systems including the learning process \cite{Goltz2021,  Sakemi2021}.

\section*{Acknowledgments}

This work was partially supported by SECOM Science and Technology Foundation, JST PRESTO Grant Number JPMJPR22C5, the NEC Corporation, Moonshot R\&D Grant Number JPMJMS2021, AMED under Grant Number JP21dm0307009, Institute of AI and Beyond of UTokyo, the International Research Center for Neurointelligence (WPI-IRCN) at The University of Tokyo Institutes for Advanced Study (UTIAS), JSPS KAKENHI Grant Number JP20H05921. Computational resource of AI Bridging Cloud Infrastructure (ABCI) provided by National Institute of Advanced Industrial Science and Technology (AIST) was used.
\bibliographystyle{IEEEtran}
\bibliography{main}

\end{document}